\documentclass[nohyperref]{article}

\usepackage{microtype}
\usepackage{graphicx}
\usepackage{booktabs} 

\usepackage{hyperref}

 \usepackage[accepted]{icml2022}

\usepackage{amsmath}
\usepackage{amssymb}
\usepackage{mathtools}
\usepackage{amsthm}
\usepackage{subfig}

\usepackage[capitalize,noabbrev]{cleveref}

\theoremstyle{plain}

\theoremstyle{definition}

\theoremstyle{remark}

\usepackage[textsize=tiny]{todonotes}

\icmltitlerunning{Adversarial Training Improves Joint Energy-Based Generative Modelling}

\begin{document}

\twocolumn[
\icmltitle{Adversarial Training Improves Joint Energy-Based Generative Modelling}




\begin{icmlauthorlist}
\icmlauthor{Rostislav Korst}{MIPT}
\icmlauthor{Arip Asadulaev}{ITMO,AIRI}
\end{icmlauthorlist}

\icmlaffiliation{MIPT}{MIPT, Moscow, Russia}
\icmlaffiliation{ITMO}{ITMO, Saint-Petersburg, Russia}
\icmlaffiliation{AIRI}{Artificial Intelligence Research Institute, Moscow, Russia}

\icmlcorrespondingauthor{Arip Asadulaev}{aripasadulaev@itmo.ru}

\icmlkeywords{Machine Learning, ICML}

\vskip 0.3in
]



\printAffiliationsAndNotice{}  

\begin{abstract}
We propose the novel framework for generative modelling using hybrid energy-based models. In our method we combine the interpretable input gradients of the robust classifier and Langevin Dynamics for sampling. Using the adversarial training we improve not only the training stability, but robustness and generative modelling of the joint energy-based models. 
\end{abstract}\vspace{-4mm}
\vspace{-5mm}\section{Introduction}
\begin{figure*}[!htb]
\centering
\minipage{0.333\textwidth}
  \includegraphics[width=0.97\linewidth]{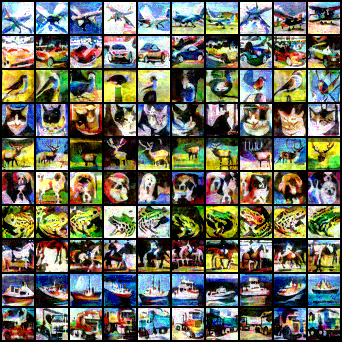}
  \caption{\label{SRC} Single Robust Classifier \\ \citep{DBLP:conf/neurips/Santurkar19}}
\endminipage
\minipage{0.333\textwidth}
\centering
  \includegraphics[width=0.97\linewidth]{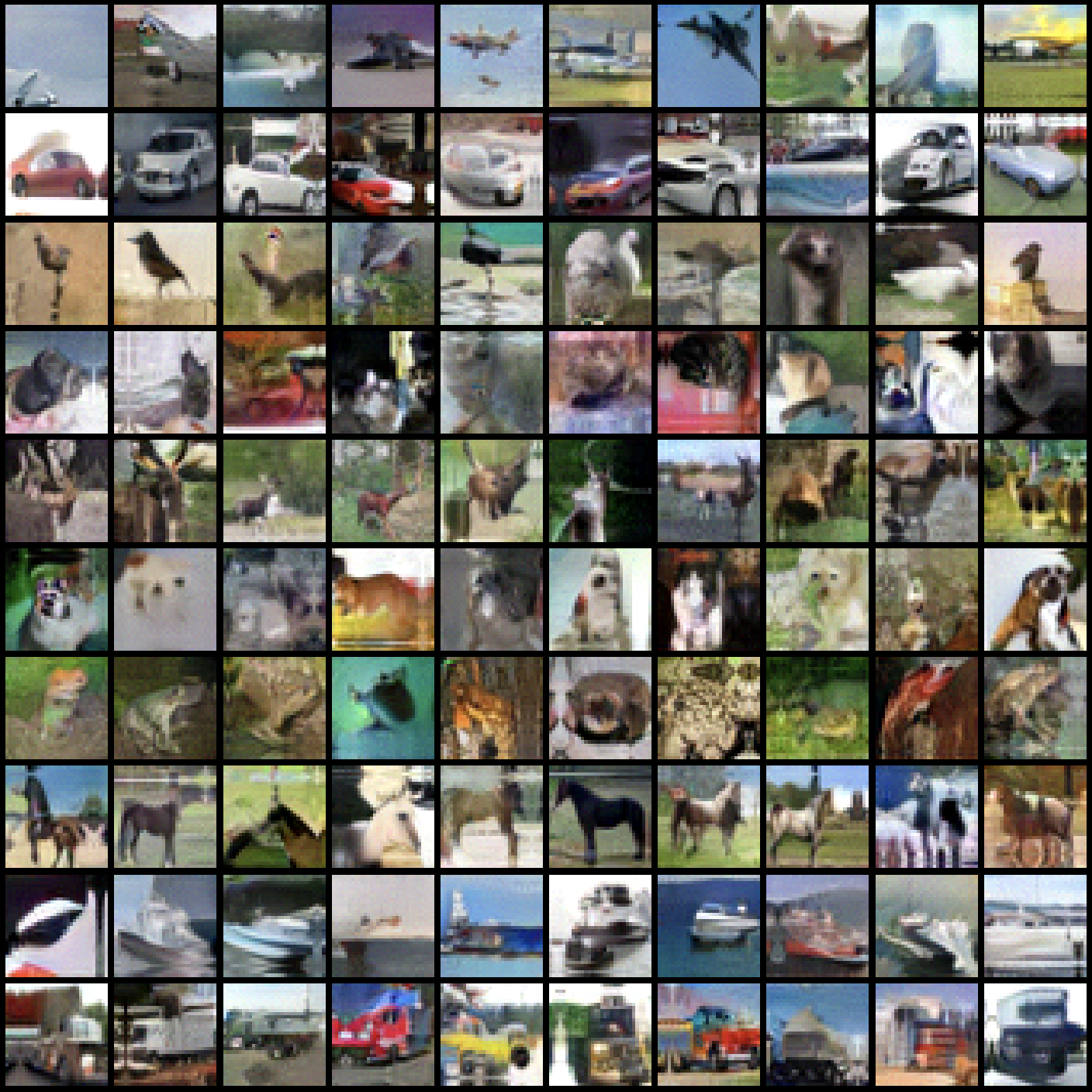}
  \caption{\label{JEM} JEM++ \\ \cite{yang2021jem++}}
\endminipage
\minipage{0.333\textwidth}%
\centering
  \includegraphics[width=0.97\linewidth]{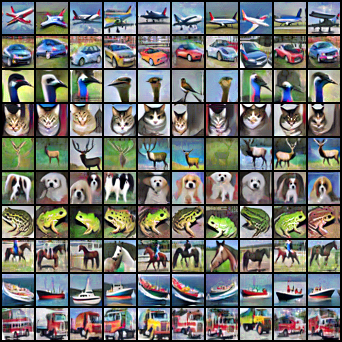}
  \caption{\label{Ours} Robust JEM++ \\ \textbf{(Ours)}}
\endminipage
\end{figure*}
\vspace{-1mm}
Adversarial examples tend classifier to make incorrect predictions.  The classifier that is not prone to adversarial examples is called adversarially robust~\cite{DBLP:journals/corr/SzegedyZSBEGF13, DBLP:conf/ccs/PapernotMGJCS17, YuanHZL19}. The robustness can be measured based on the visual similarity of adversarial attack gradients to the real data~\cite{DBLP:journals/corr/abs-1905-08232}. Moreover, when classifier is robust, adversarial attack adds robust features to the image, that is visually similar to the data~\cite{DBLP:journals/corr/abs-1905-02175}. Based on this, robust classifiers can be used for images generation~\cite{DBLP:conf/neurips/Santurkar19}. 

In this paper we combine robust classifier abilities with the energy-based models for generative modelling. It was shown that classifier can be reinterpreted as an energy-based model for the joint distribution~\cite{grathwohl2019your}.  During image generation, (1) we sample the gaussian noise, (2) then turn samples into specific class using the adversarial attack, (3) and then minimize the energy using Stochastic Gradient Langevin Dynamics (SGLD).  

Adversarial training can improve the out-of distribution detection of the energy based models~\cite{grathwohl2019your, lee2020adversarial}, but to best of our knowledge, the connection between adversarial training and energy-based inference for generative modelling was not previously studied. We tested our method on CIFAR-10 dataset and improved IS and FID in comparison to the standard robust classifier and joint energy-based model. 
\section{Background}
\textbf{Joint Energy-Based Model:} Energy-based model (EBM) learns to assign low energy values to samples
from training distribution and high values otherwise. The probability density $p_{\theta}(x)$ given by an EBM can be presented as: $p_{\theta} (x) = \frac{exp(-E_{\theta}(x))}{Z(\theta)}$,
where $E_{\theta}(x)$ is an energy function that maps each input $x$ to a scalar, and $Z(\theta)$ is an intractable normalizing constant. To generate samples by energy function, Stochastic Gradient Langevin Dynamics (SGLD) \cite{welling2011bayesian} can be used:
\begin{equation}
    x_{i+1} \leftarrow x_{i} - \frac{\alpha}{2}\frac{\partial E_{\theta}(x_i)}{\partial x} + \beta, \quad \beta \sim \mathcal{N}(0, \alpha)
\end{equation}
Recently Joint Energy-based Model (JEM) was proposed~\cite{grathwohl2019your}.  JEM reinterprets a standard discriminative classifier of $p(y|x)$ as an energy-based model for the joint distribution $p(x, y)$. This model can classify images and generate realistic samples within one hybrid model. In our experiments we used JEM++~\cite{yang2021jem++} that is more stable and fast version of JEM.

\textbf{Adversarial Examples:} Having the sample $x$, the real label $y_{real}$, model with parameters $\theta$ and loss function $L$, we can apply Projected Gradient Descent  (PGD)~\cite{MadryMSTV18} to generate the adversarial examples:
\vspace{-1mm}
\begin{equation}
    {x}_{i+1}=\operatorname{Proj}_{(x, \varepsilon)}\left[{x}_{i}+\alpha \operatorname{sign}\left(\nabla_{x} L\left(\theta, {x}_{i}, y_{real}\right)\right)\right]
\end{equation} 
Where, $\operatorname{Proj}_{(x, \varepsilon)}$  is a projection operator onto the $l_{\inf}$  ball of radius $\varepsilon$ around the original image $x_i$. After obtaining adversarial examples we can train the model on this examples to increase the robustness~\cite{DBLP:journals/corr/SzegedyZSBEGF13}. The robust classifier can generate realistic images by minimizing cross-entropy loss on target class with PGD that is called target attack \cite{DBLP:conf/neurips/Santurkar19}. 

\section{Method}
We propose to combine adversarial training of classifier and joint energy-based model. During training process we fed classifier with adversarial examples and we find that it improves training stability of JEM. Due to the use of adversarial training, the classifier contained in hybrid model got adversarially robust and obtained the ability to generate images with PGD attack~\cite{DBLP:conf/neurips/Santurkar19}. Moreover, our hybrid model can generate images by minimizing energy value similar to baseline JEM. We propose to combine both generation methods into \textbf{combined inference}. Energy-based SGLD steps can be treated as perturbation smoothing mechanism for prior obtained with target attack on noise. 



To perform conditional generation we start from Gaussian mixture distribution estimated from the training dataset like in \cite{yang2021jem++}, then we implement target PGD attack with classifier contained in JEM. Finally we minimize joint energy function using Langevin dynamics given joint energy function $E_{\theta}(x, y) = - f_{\theta} (x) [y]$, proposed in \cite{grathwohl2019your}, where $f_{\theta}(x)$ - is a parametric function $\mathbb{R}^D \rightarrow \mathbb{R}^K$, which maps each data point $x \in \mathbb{R}^D$ to $K$ real-valued numbers known as classifier logits and $f_{\theta} (x) [y]$ - the logit corresponding to the $y^{th}$ class label. In other words \textbf{the initial sampling distribution for JEM is defined by adversarial attack to the classifier contained in JEM}.
\section{Experiments}
\label{Experiments}
\textbf{Dataset:} We trained and evaluated our models on CIFAR-10 dataset. The CIFAR-10 dataset consists of 60000 32x32 colour images in 10 classes, with 6000 images per class. 

\textbf{Settings:} We used JEM++ \cite{yang2021jem++} for our experiments. All architectures used are based on Wide Residual Networks~\cite{zagoruyko2016wide}. JEM++ uses $m$
outer loops and $n$ inner loops for SGLD steps. For training JEM++ we use $m=10$, $n=5$ steps, and $n=300$, $n=5$ for unconditional inference. Same parameters were used for our model. For conditional inference we use $m=50$, $n=5$. Adversarial attack parameters for adversarial training was: $\varepsilon_{adv}=0.1$, number of steps is 15, constraint is $inf$. For inference we use $\varepsilon_{adv}=0.5$. Same parameters were used for training single robust classifier. In combined inference, we decrease contrast of adversarial prior, it allows pixels not to go beyond the range of $[-1, 1]$ after SGLD and not to clamp pixels. Empirically, this trick improves the performance.

\textbf{Metrics:} To evaluate the quality of generated images, we used the Inception Score (IS)~\cite{salimans2016improved} and Frechet Inception Distance (FID)~\cite{heusel2017gans}.
\begin{table}[h!]
\centering
\begin{center}
\begin{small}
\begin{tabular}{l|ll}
\hline
Model      & IS $\uparrow$    & FID$\downarrow$ \\ \hline
Residual Flow~\cite{chen2019residual}      & 3.69  & 46.40 \\
Glow~\cite{kingma2018glow}      & 3.92  & 48.90 \\
JEM++~\cite{yang2021jem++}     & 7.54  & 48.25 \\
Single Robust Classifier~\citep{DBLP:conf/neurips/Santurkar19}        & 7.05  & 85.12 \\ \hline
Robust JEM-Energy (Ours) & 8.71  & \textbf{41.17} \\ 
Robust JEM (Ours) & \textbf{9.28}  & 47.92 \\      
\hline
\end{tabular}
\end{small}
\end{center}
\vskip -0.1in
\caption{Hybrid models results on CIFAR-10. Robust JEM-Energy is sampling using only SGLD steps without adversarial attack. Samples were generated from scratch.}
\label{tab:table-1}
\end{table}
\vskip -0.2in
\textbf{Results:} We compared generative performance of our model (Figure~\ref{Ours}) with JEM++(Figure~\ref{JEM}) and Single Robust Classifier~\citep{DBLP:conf/neurips/Santurkar19},(Figure~\ref{SRC}). Our model improved performance of energy-based inference of JEM++, see Table~(\ref{tab:table-1}). The combined inference showed increase in IS rivaling the hybrid models state-of-the-art in generative learning.

\section{Conclusion and Future Work}
Thanks to the strong adversarial training, JEM improves its generative performance from noise. The study of visually interpretable gradients of robust networks is still unrepresented research area. In our opinion this property can find many applications in generative modelling. As a problem, we find that our model tend to generate less diverse images than standard energy based models and we are going to tackle this issue in the future work. 


\newpage

\bibliography{example_paper}
\bibliographystyle{icml2022}

\end{document}